\pdfoutput=1

\documentclass[11pt]{article}

\usepackage[]{acl}

\usepackage{times}
\usepackage{latexsym}

\usepackage[T1]{fontenc}

\usepackage[utf8]{inputenc}

\usepackage{microtype}

\usepackage{algorithm}
\usepackage{algorithmic}

\usepackage{newfloat}
\usepackage{listings}
\floatstyle{ruled}
\newfloat{listing}{tb}{lst}{}
\floatname{listing}{Listing}

\usepackage[utf8]{inputenc}
\usepackage{times}
\usepackage{latexsym}
\usepackage{graphicx}
\usepackage{graphics}
\usepackage{enumitem}
\usepackage{tablefootnote}
\usepackage{multirow}
\usepackage{xcolor}
\definecolor{maroon}{rgb}{0.6, 0.0, 0.0}

\newcommand\ruletaker{\textsc{RuleTakers}}
\newcommand\lot{\textsc{Leap-Of-Thought}}
\newcommand\clutrr{\textsc{CLUTRR}}

\title{Reasoning is about giving reasons}

\author{Krunal Shah\footnotemark[1] \\
  Yutori Inc. \\
  \texttt{ktgshah@gmail.com} \\\And
  Dan Roth \\
  University of Pennsylvania \\
  \texttt{danroth@seas.upenn.edu} \\}

\begin{document}

{\makeatletter\acl@finalcopytrue
  \maketitle
}

\renewcommand{\thefootnote}{\fnsymbol{footnote}} 
\footnotetext[1]{Work done while the author was a student at the University of Pennsylvania.}
\renewcommand{\thefootnote}{\arabic{footnote}} 
\begin{abstract}

Convincing someone of the truth value of a premise requires understanding and articulating the core logical structure of the argument which proves or disproves the premise. Understanding the logical structure of an argument refers to understanding the underlying ``reasons" which make up the proof or disproof of the premise - as a function of the ``logical atoms” in the argument.
While it has been shown that transformers can ``chain” rules to derive simple arguments, the challenge of articulating the ``reasons” remains. Not only do current approaches to chaining rules suffer in terms of their interpretability, they are also quite constrained in their ability to accommodate extensions to theoretically equivalent reasoning tasks – a model trained to chain rules cannot support abduction or identify contradictions.

In this work we suggest addressing these shortcomings by identifying an intermediate representation (which we call the Representation of the Logical Structure (RLS) of the argument) that possesses an understanding of the logical structure of a natural language argument -- the logical atoms in the argument and the rules incorporating them. 
Given the logical structure, reasoning\footnote{This work focuses on extracting the propositional structure of the arguments, and does not attempt to handle quantifiers, as is the case for most recent neural work on reasoning.} is deterministic and easy to compute.
Therefore, our approach supports all forms of reasoning that depend on the logical structure
of the natural language argument, including  arbitrary depths of reasoning, on-the-fly mistake rectification and interactive discussion with respect to an argument. We show that we can identify and extract the logical structure of natural language arguments in three popular reasoning datasets with high accuracies, thus supporting 
{\em explanation generation} and extending the reasoning capabilities significantly.\footnote{This work was originally written in 2021 and was never published. It suggests that Language Models could support reasoning and interpretability better by converting text to an intermediate formal representation that solvers can reason about deterministically. While the Language Models have improved significantly since then, this line of thought is still valid and very important to pursue.}

\end{abstract}

\section{Introduction}

\begin{figure}
\centering
\includegraphics[width=7.6cm,height=4.5cm]{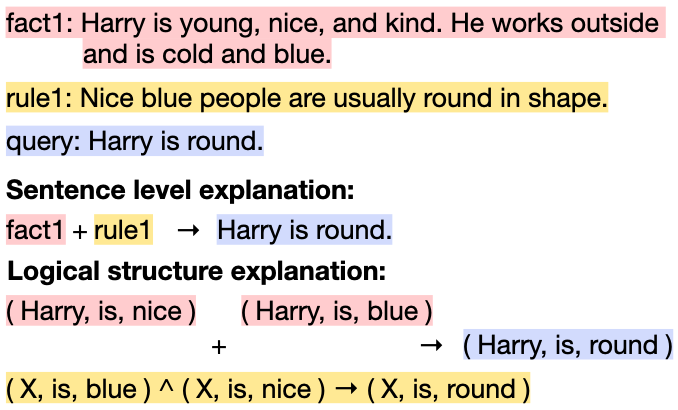}
\caption{\label{fig:explain} An 
example of an {\em argument} from the \ruletaker{} \cite{ruletaker} dataset. The goal is to determine the truth value 
of the query given some facts and rules (that all together constitute an argument). The logical structure of the argument at the bottom
conveys a deeper understanding of the reasoning process, simplifies the verification process, and supports {\em reasoning} on ``What if Harry was young but not nice?" and "If Harry is round then does that make him nice?".
}
\end{figure}

Natural language is often used to make arguments. For example, given some facts and rules, a deductive argument can be made, as in Fig.~\ref{fig:explain}.
For many years, the key approach in AI, to reasoning argued in natural language (NL) text, was to map the text to a formal structure and then use well defined, deterministic, algorithms that support exact reasoning over these representations~\cite{Hayes1977InDO,HSME88,McCarthy63,McCarthy76}.
However, latter works have discussed the brittleness of this approach, arguing that, in general, natural language text is too expressive to be mapped into a formal, symbolic representation~\cite{Charniak1973JackAJ} 


The recent success of neural models and, more specifically, transformer based language models, has prompted researchers to explore the reasoning abilities of such systems. Recent work \cite{ruletaker, saha-etal-2020-prover, proofwriter} demonstrates the ability of transformers to perform simple chaining of rules (top half of Fig.~\ref{fig:explain}) and some limited form of abduction \cite{proofwriter}.

However, it is important to note that the only level of uncertainty that exists in reasoning problems expressed in natural language
is the derivation of 
the core logical structure of the underlying NL arguments. Once the logical structure of an argument is given, the computational problem is deterministic and easy; it boils down to solving satisfiability or unsatisfiability over a small domain.

In this paper we first agree with previous works that in general, it is not possible to map natural language to a symbolic representation; however, we argue that it is possible to map \emph{arguments} made in natural language to their core logical structure (where this mapping might lose some information that is not relevant to the argument itself).
Towards this end, we define a class of structured representations, in the context of reasoning tasks, called Representation of Logical Structure (RLS). The primary contribution of this work is defining the RLS of an argument made in text and showing how, by doing so,
we can {\em explain} the argument and support the benefits of performing symbolic reasoning over knowledge expressed in natural language. 
Extracting the logical structure of an argument in the form of structured representations (RLS) allows us to {\em explain} the reasoning process, support other forms of reasoning that depend on this logical structure, and {\em reason} with respect to it -- interact about it, identify fallacies, etc. These latter steps can be done by simply feeding the extracted representations to an appropriate symbolic reasoning engine. This is a well defined and solved problem ~\cite{Gu1996AlgorithmsFT} and we do not attend to this part of the computation in this paper.

Specifically, we propose using a 
sequence to sequence transformer to extract RLS from natural language text. We demonstrate the effectiveness of our RLS extraction method, in capturing the logical structure of natural language arguments, on three datasets spanning a variety of reasoning tasks and language usage.
First, our RLS extraction method shows its success on the \lot{} \cite{lot} and \clutrr{} \cite{sinha-etal-2019-clutrr} datasets by registering 99\% and 95\% exact match accuracies, showing that the representation extraction method can have broad coverage.
Furthermore, we show the effectiveness of using a symbolic reasoner over the extracted logical structures on the data
that was used in \cite{ruletaker, saha-etal-2020-prover, proofwriter} and demonstrate the effectiveness of this approach.
Specifically, we find that our method performs competitively 
with the current end-to-end reasoning systems on the rule based reasoning task of the \ruletaker{} \cite{ruletaker} benchmark. 

In the following section, we make a more detailed case for our methodology by addressing some of the challenges of introducing intermediate logical representations, 
along with the benefits of this approach. 
Finally, we note that the primary motivation for this work is not to improve the performance on existing benchmarks but to (1) emphasize the need for finer logical explanations for reasoning tasks, (2) investigate the effectiveness of existing language models in extracting structured logical representations\footnote{we use ``structured representations", ``structured logical representations" and ``logical representations" interchangeably} from arguments expressed in natural language, and (3) throw light on our alternate, more symbolic approach which leverages the language modeling advances to perform competitive, more explainable reasoning compared to the end to end approaches adopted by previous works.

\section{``Explanations" are symbolic}
\label{sec:explanations}
In this section we try to lay out a more detailed analysis addressing some of the shortcomings of the ``extracting structured logical representations" approach and subsequently outline the notable advantages of this approach. 
\subsection{Challenges}
\label{challenges}

An important shortcoming of using symbolic reasoning methods which, as discussed earlier, was the key conceptual barrier to early efforts, is that natural language is simply too expressive. Given the diverse linguistic variability of natural language, and the amount of information humans are able to communicate via natural language, we cannot expect to accurately and reliably convert NL utterances into structured representations. 
This paper agrees with this assessment and does {\em not} attempt to convert NL to a structured representation. Instead, our method of extracting the RLS of 
NL arguments 
relies 
on a much weaker assumption about the presence of a ``local" logical structure in NL arguments. 
And in extracting the logical structure we admittedly lose some other aspects of the text, that are not relevant to the logical argument (but could very well be relevant to other decisions made with respect to the text). For example, if Figure~\ref{fig:explain} were to say that ``Harry works outside, to his chagrin" then this will not affect the logical structure of the argument, and will not impact the logical atoms extracted by the method, since the mental state of Harry, while important, is irrelevant to the logical argument.


Another notable challenge of using symbolic reasoning methods with natural language is associated with their inability to handle the noise, uncertainty, and ``softness" expressed in language. However, recent work \cite{lewis-steedman-2013-combined,weber-etal-2019-nlprolog} has shown promising results in trying to mitigate these by making use of weak unification operators to replace the exact matching in symbolic reasoners. 
When this ``soft" unification is done by a symbolic solver, the  solver provides insights into the assumptions made during the reasoning process as shown in Figure~\ref{fig:unify}, since the logical atoms which are considered equivalent by the unification operator represent the linguistic/implicit knowledge encoded in the unification operator function.
That is, if you know that the reason the outcome is such is due to unifying the variables ``has gold" and ``has a heart of gold", the process becomes more transparent.
\begin{figure}
\begin{center}
\begin{tabular}{p{0.98\linewidth}}
\emph{fact:} Mary is a young woman with a heart of gold. \\
\emph{rule:} If Mary has gold then she is rich.
\end{tabular}
\end{center}
\caption{\label{fig:unify} In the above inference, if the weak unification operator concludes that ``(Mary, has, heart of gold)" and ``(Mary, has, gold)" are equivalent then the symbolic reasoner would explicitly indicate the same in the proof, providing insights into the implicit assumptions/knowledge used in the reasoning process.
}
\end{figure}

We note that
in this work, we only perform extraction of structured representations from individual natural language utterances rather than from complete (possibly multi-sentence) arguments. 
We leave the problem of extending the extraction to account for the whole context of the natural language argument for future work.

\subsection{Symbolic reasoning is ``simple"}
\label{sec:simple}
In this subsection, we draw attention to the observation that the primary challenge in the problem of reasoning over natural language text is concerned with the ambiguity and linguistic variability associated with natural language. Once we can accurately map the relevant information from the NL text into its structured logical representations, the task of reasoning over these symbols is nearly trivial. Propositional symbolic reasoning is well understood, and essentially boils down to boolean satisfiability or unsatisfiability; this can be performed very efficiently~\cite{Gu1996AlgorithmsFT,Gomes2008SatisfiabilityS} over a small number of variables.



\subsection{Advantages}
\label{sec:advantages}
The foundations of reasoning laid out in the field of artificial intelligence were primarily defined to deal with symbolic reasoning, as a result, there is rich and diverse literature around symbolic reasoning. Our approach of extracting the logical representations from NL sentences allows us to take full advantage of these advances. 
We now try to summarize some of the advantages of our system which uses a symbolic reasoning engine on top of extracted logical structures. 
\begin{enumerate}[leftmargin=*,itemsep=3pt]
    \item Note that once we can accurately extract the representations from the NL sentences, we can perform any kind of reasoning (e.g. deductive, abductive, contradiction detection, etc.) that depends on this logical structure, without the need for any additional training. This is in contrast to the existing end-to-end transformer based systems which would require additional training data to learn any new kind of reasoning.
    \item As discussed in Section \ref{challenges}, the usage of weak unification operators provide a potential solution to addressing the inability of symbolic reasoning systems to handle the ``softness", noise and ambiguity expressed in natural language. We observe that the usage of language models as weak unification operators additionally allows us to (1) leverage the linguistic and implicit knowledge of the language model and (2) explicitly lay out the unifications used in inference, as demonstrated in Figure~\ref{fig:unify}.
    \item Given the disentanglement of the reasoning responsibilities in our method, the system allows for online mistake rectification and tweaks. For example, if a mistake is observed in the representation extraction step during inference, the mistake can be rectified by human intervention and the resulting corrected theory can be simply fed through the symbolic reasoning engine to get the correct answer. In contrast, the end-to-end systems do not provide any outlet to rectify incorrect inferences drawn by the model.
    \item The above mentioned disentanglement also allows us to transfer the method to new kinds of rules or facts (like multivariate rules and event attributes based reasoning) by simply training a new representation extraction model. Training the end-to-end systems on the other hand requires additionally gathering complete theories of reasoning over the natural language sentences.
\end{enumerate}


\section{RLS and problem formulation}
In this section, we define the structured representations proposed in this work, termed Represention of Logical Structure (RLS). We also detail how we formulate RLS for the datasets used for our experiments and then briefly describe how we formulate the problem of extracting RLS from natural language text.

\subsection{Representation of Logical Structure (RLS)}
We define RLS for a natural language statement $S$, in the context of a reasoning task $R$, as a structured representation which encodes all the relevant information in statement $S$, i.e. for any inference/reasoning chain which $S$ is a part of, if we replace the statement $S$ with its corresponding RLS then the resulting inference/reasoning chain still remains valid and evaluates to the same result as with $S$. Although not included in the definition, a desirable property from a good RLS formulation is that it is efficient, i.e. it only encodes the information from statements that is truly relevant to the reasoning task. More specifically, if an alternate RLS formulation for the given task $T$ encodes strictly less information than the existing formulation while still satisfying the definition of RLS then the existing formulation would not be called efficient. Kindly note that we propose the RLS of a natural language statement as an alternative to the semantic parse of the statement. Hence, similar to the semantic parse of a statement, RLS is (a) task dependent, (b) a class of structured representations and (c) does not follow a well-defined syntax. 

Now we try to briefly emphasize the differences of our approach of extracting RLS compared to semantic parsing. Unlike semantic parsing which aims to extract structured logical representations from \emph{all forms} of natural language text with \emph{minimal information loss}, extracting RLS is only defined in the context of a reasoning task meaning that it aims to only extract structured logical representations from NL sentences where a logical structure can be expected to exist. Furthermore, RLS only aims to extract information from a sentence which is essential to the usage of the sentence in the context of the reasoning task, hence extracting RLS has a looser requirement on information loss compared to semantic parsing. We posit that these fundamental differences help us mitigate some of the primary issues with semantic parsing that originate from the diverse variability of natural language and the amount of information humans are able to communicate via language. 

\subsection{Our RLS formulation}
We now briefly describe the RLS formulations used by our approach to encode the logical structure from natural language statements, for the problems used in our experiments. Our structured representations are simple logical statements where the logical literals are tuples of natural language words, phrases and binary symbols. Table~\ref{table:representations} shows some examples of natural language sentences and the corresponding RLS used by our approach. Since RLS is defined in the context of a particular reasoning task, note that the representations follow different formulations for the different datasets and the choice for the formulation is primarily dictated by the kind of logical structure that is important for the corresponding reasoning task. 

\subsection{Problem formulations}
\subsubsection{Representation extraction formulation} We model the problem of extracting RLS from a natural language sentence as a sequence to sequence prediction task where the input is the natural language utterance and the output is an encoding of the RLS of the NL sentence. Table~\ref{table:representations} shows some examples of how the structured representations from instances of the different datasets are encoded for the sequence output of the model.

\subsubsection{Rule based reasoning} The deductive reasoning task described in \citealt{ruletaker} takes as input a natural language theory, consisting of a set of rules and facts described in natural language, and a query described in NL. The goal is to predict the truth value of the query, in the world defined by the theory, under the closed world assumption (CWA).

\section{Extracting the Logical Structure of Arguments
}
\label{sec:experiments}
\begin{table*}
\centering
\resizebox{\linewidth}{!}{%
\begin{tabular}{l|l|l}
\hline
Dataset           & Type           & Sentence                                                                                                                                                                                                                                                                                     \\ \hline
\ruletaker{} (fact) & Sentence       & Harry is young and nice.                                                                                                                                                                                                                                                                     \\
                  & Annotation & ("Harry" "is" "young" "+")  ("Harry" "is" "nice" "+")      \\
                  & RLS & (Harry, is, young, +) $\wedge{}$ (Harry, is, nice, +)      \\
                  & Encoded Repr. & \textless{}arg0\textgreater\ Harry \textless{}pred\textgreater\ is \textless{}arg1\textgreater\ young \textless{}pos\textgreater\ \textless{}and\textgreater\ \textless{}arg0\textgreater\ Harry \textless{}pred\textgreater\ is \textless{}arg1\textgreater\ nice \textless{}pos\textgreater{}      \\ \hline
\ruletaker{} (rule) & Sentence       & Nice people are usually round in shape.                                                                                                                                                                                                                                                      \\
                  & Annotation & ("someone" "is" "nice" "+") -\textgreater{} ("someone" "is" "round" "+") \\
                  & RLS & (someone, is, nice, +) -\textgreater{} (someone, is, round, +) \\
                  & Encoded Repr. & \textless{}arg0\textgreater\ someone \textless{}pred\textgreater\ is \textless{}arg1\textgreater\ nice \textless{}pos\textgreater\ \textless{}impl\textgreater\ \textless{}arg0\textgreater\ someone \textless{}pred\textgreater\ is \textless{}arg1\textgreater\ round \textless{}pos\textgreater{} \\ \hline
\clutrr{}            & Sentence       & Sol took her son Kent to the park for the afternoon.                                                                                                                                                                                                                                         \\
                  & Annotation & edges: [("Sol", "Kent")], edge\_types: ["son"], genders: \{"Sol": female, "Kent": male\}                                                              \\
                  & RLS & (Sol, son, Kent) $\wedge{}$ (Kent, mother, Sol)                                                              \\
                  & Encoded Repr. & \textless{}arg1\textgreater\ Sol \textless{}pred\textgreater\ son \textless{}arg2\textgreater\ Kent \textless{}and\textgreater\ \textless{}arg1\textgreater\ Kent \textless{}pred\textgreater\ mother \textless{}arg2\textgreater\ Sol                                                              \\ \hline
\lot{}   & Sentence       & A mustard is not capable of shade from sun.                                                                                                                                                                                                                                                  \\
                  & Annotation & \{"subject": "mustard", "predicate": "/r/CapableOf", "object": "shade from sun", "validity": "never true"\}  \\
                  & RLS & (mustard, capable of, shade from sun, -)  \\
                  & Encoded Repr. & \textless{}arg0\textgreater\ mustard \textless{}pred\textgreater\ capable of \textless{}arg1\textgreater\ shade from sun \textless{}neg\textgreater{}  \\    \hline                                                                                                                                     
\end{tabular}
}
\caption{\label{table:representations} Table showing how the annotations provided with the datasets are converted to their respective RLS formulations and subsequently encoded to train the sequence to sequence representation extraction model.}
\end{table*}
In this section, we discuss the questions we wish to answer in our experiments and briefly describe the datasets, models and evaluation metrics used for the same.

\subsection{Experimental settings}
\label{sec:questions}
The goal of our experimental evaluation is two fold, to first investigate whether our RLS extraction method works well for different kinds of RLS formulations and natural language text; and subsequently understand the efficacy of our reasoning system which relies on the RLS extraction method.
Specifically, we design experimental settings to answer the following questions:
\begin{enumerate}
    \item Can the RLS extraction method generalize well to different kinds of logical reasoning datasets which can benefit from reasoning over structured logical representations instead of natural language text?
    \item How well can a method relying on RLS extraction perform on the rule based reasoning task (example in Fig.~\ref{fig:explain}), compared to the end-to-end transformer systems?
\end{enumerate}

We use the \lot{} \cite{lot} and \clutrr{} \cite{sinha-etal-2019-clutrr} datasets to test the generalizability of our representation extraction method to answer our first question; and further use the \ruletaker{} dataset \cite{ruletaker} as the benchmark for the rule based reasoning task to answer the second question. We now briefly describe each of these datasets.


\subsection{Datasets} 

\begin{enumerate}[topsep=3pt,leftmargin=*,itemsep=3pt]
    \item \textbf{\lot{}} \cite{lot}: A dataset designed to test the ability of models to reason over implicit factual knowledge. Instances of the dataset include a collection of facts and rules, expressed in templated language, along with a query such that answering the query requires some implicit factual knowledge along with the explicitly stated facts and rules. This dataset is used to test whether our representation extraction method can handle RLS representations which primarily consist of phrases (as opposed to simple words) as the logical literals' arguments (as shown in Table~\ref{table:representations}).
    For our experiments, we merge all the different train and test sets corresponding to the two settings of ``counting" and ``hypernymy" which results in around 38k/22k unique training instances and 4811/3234 unique test instances for the hypernymy/counting settings respectively and we train separate representation extraction models for the two settings. The logical literals for the statements in this dataset follow the pattern $(noun\_phraseA, relation, $ $noun\_phraseB, polarity)$. The logical literals combine by using simple logical conjunction operators to form the RLS for the natural language statements.
    \item \textbf{\clutrr{}}~\cite{sinha-etal-2019-clutrr}: An inductive reasoning benchmark which requires a model to infer a kinship relation that is not explicitly stated in the input while also learning the logical rules governing the relations. The benchmark consists of a natural language setting where the input relations are expressed using NL and a graph based setting where the relations are provided as symbolic inputs. The natural language descriptions are curated using crowdsourced templates and the test set consists of unseen NL templates. We use this dataset to test our representation extraction method's ability to generalize to more natural forms of language and to a different kind of RLS formulation. Note that since the dataset deals with reasoning over kinship relations, we identify the kinship relations between entities as the logical structure to extract from instances of the dataset, as shown in Table~\ref{table:representations}. For our purpose, we generate a dataset of 8k training instances and 846 test instances of unseen templates. The logical literals for the statements in this dataset follow the pattern $(personA, relationAB, personB)$. The logical literals combine by using simple logical conjunction operators to form the RLS for the natural language statements. Note that since the inverse of kinship relations exist, for every relation $(personA, relationAB, personB)$, we also include the inverse relation literal $(personB, relationBA, personA)$ in the RLS for the given natural language statement.
    \item \textbf{\ruletaker{}}~\cite{ruletaker}: A collection of several rule based reasoning datasets where the instances consist of theories and related queries expressed in natural language. The theories consist of facts and rules about entity relations and attributes and the queries require deductive reasoning over the theory of facts and rules (example in Fig.~\ref{fig:explain}). We specifically use the D3, D5, ParaRules and Birds-Electricity datasets for our experiments to investigate the performance of our reasoning method which uses a symbolic reasoner on top of the RLS extraction model. The D3 and D5 datasets contain 100k questions where the theories and queries are expressed in templated natural language and the queries require deductive reasoning up to 3 and 5 depths respectively. The ParaRules dataset contains 40k questions where the theories are paraphrased versions of the templated theories. The paraphrased templates were collected using crowdsourcing and the dataset is designed to test whether the methods can generalize to more natural forms of language. Birds-Electricity is a collection of several hand authored rulebases where the Birds rulebases are adapted from ``birds" logic problem from \citealt{10.1016/0004-3702(86)90032-9} and Electricity rulebases were generated by \citealt{ruletaker}. The dataset is used for zero-shot evaluation, to test the out-of-domain generalization ability of the models since the language, entities and attributes in the dataset are different from those observed in the training data. The logical literals for the statements in these datasets follow the pattern of either $(entity, relation, entity, polarity)$ or $(subject, is, property, polarity)$. The logical literals combine by using simple logical conjunction operators to form the RLS for the factual statements. The logical atoms combine in the form $l_1 [\wedge{}\ l_i]^* \rightarrow l_c$ to form RLS for the rule statements in the datasets.
\end{enumerate}
We note here that for all of the above datasets, we are able to easily generate the RLS representations from the accompanying metadata available with the datasets (as described in Table~\ref{table:representations}) and hence no annotation effort was required in this work.

\subsection{Models and Evaluation}
\label{sec:models}
\paragraph{Baseline} For the \ruletaker{} dataset, the state of the art ProofWriter \cite{proofwriter} and the earlier PRover \cite{saha-etal-2020-prover} models are used as the baselines methods. For our experiments on the \clutrr{} and \lot{} datasets we do not have any baseline methods because we wish to show strong performances on the representation extraction task and do not try to solve the respective end tasks.

\paragraph{Our models}
We use a pretrained text to text transformer model, namely T5 \cite{t5}, for all our experiments to extract structured representations from input NL text. More specifically, the model takes as input a natural language sentence and learns to predict the encoded RLS representation of the input sentence, where the encoding of the RLS representation is performed as shown in Table~\ref{table:representations}. We use a T5-small model to train on the D3 dataset while all other models are trained using a T5-base model. The model is trained to predict a structured representation of the relevant information from the input utterances. The structured representations for the different datasets are encoded as shown in Table~\ref{table:representations}. 
For experiments on the \ruletaker{} dataset, we make use of ProbLog \cite{10.5555/1625275.1625673} to perform inference on the extracted logical representations.

\paragraph{Evaluation} For the representation extraction task for the \lot{} and \clutrr{} datasets, we use exact matching accuracy on the extracted logical representations as the evaluation metric. Also note that we process the predicted sequence outputs assuming they follow the expected representation encoding protocol described in Table~\ref{table:representations}. However, predictions which do not adhere to the same would result in incorrect exact matching for the representation extraction metric or result in incorrect predictions on the downstream reasoning task.
For the deductive reasoning tasks on the \ruletaker{} dataset, where the goal is to predict the truthfulness of the query sentence given explicit rules and facts, we use the answer accuracy as the evaluation metric. Note that we do not report proof accuracies which are previously reported in \cite{proofwriter, saha-etal-2020-prover} because the metric makes sense for the end-to-end transformer systems to demonstrate that the models are predicting correct answers for the ``right" reasons; however, in our case since we use a symbolic reasoning engine it is clear that our predictions are completely faithful to the explanations generated by our method. 
\section{Results}
\label{sec:results}
In this section we analyse the results of our method on various experimental settings, to answer the questions we ask in Section~\ref{sec:questions} about the extensibility of our representation extraction model and the effectiveness of our natural language reasoning method.

\subsection{Representation extraction}
In this subsection we are interested in the generalization ability of the representation extraction model to different forms of structured representations and natural language. Table~\ref{table:clutrr} shows the performance of the representation extraction model on the \clutrr{} and \lot{} datasets and the high performance (95\%+ and 99\%+ exact match accuracy respectively) of the model demonstrates its ability to generalize to natural forms of language and different kinds of structured representations. 
We note here that even though solving the end task is not our objective in this experimental setting, as discussed earlier, there is notable merit in reducing these tasks to reasoning over the structured representations over input NL utterances as shown in \citealt{sinha-etal-2019-clutrr}, where the graph based methods exhibit much better generalization compared to the NL based methods.

\begin{table}[t]
\centering
\resizebox{\columnwidth}{!}{%
\begin{tabular}{c|c|cc}
\hline
\multirow{2}{*}{}                  & \multirow{2}{*}{\clutrr{}} & \multicolumn{2}{c}{\lot{}}       \\ \cline{3-4} 
                                   &                         & \multicolumn{1}{c|}{Hypernyms} & Counting \\ \hline
EM Accuracy & 95.9                    & \multicolumn{1}{c|}{99.6}      & 99.8  \\ \hline  
\end{tabular}
}
\caption{\label{table:clutrr} Results of the representation extraction model trained and tested on the \clutrr{} and \lot{} datasets, where we report the exact match accuracy of the extracted structured representations. The high accuracy scores show the models' generalization ability to different kinds of natural language text and structured representations.}
\end{table}

\subsection{Rule based reasoning}
In this subsection we investigate if our method which relies on RLS as intermediate logical representations can perform competitively to the state of the art baseline systems on the rule based reasoning task.

\subsubsection{ParaRules dataset}
\begin{table}[t]
\centering
\begin{tabular}{c|c|ccc}
\hline
D & \#qns & Ours\footnotemark[4] & ProofWriter & PRover \\ \hline
0         & 2968  & 99.6       & 99.9        & 99.7   \\
1         & 2406  & 99.2       & 99.3        & 98.6   \\
2         & 1443  & 96.4       & 98.3        & 98.2   \\
3         & 1036  & 93.1       & 98.2        & 96.5   \\
4         & 142   & 87.3       & 91.5        & 88.0   \\ \hline
All       & 8008  & 97.8       & 99.1        & 98.4   \\ \hline
\end{tabular}
\caption{\label{table:pararules} Results of systems trained on D3 + ParaRules and tested on ParaRules where our system demonstrates competitive performance to the current SToA systems.
}
\end{table}
\footnotetext[4]{results on the subset of the test set where ProbLog did not throw an exception on the corresponding gold theories. This is an error in the dataset as pointed out in \citealt{proofwriter}.}
Firstly, to show that our method can perform competitively to the state of the art models on naturally occuring text, we train a model on the combined sentences from D3 + ParaRules and test it on the ParaRules dataset in Table~\ref{table:pararules}. We observe that our method performs competitively (within 1.3\%) to the baseline methods thus demonstrating the overall strong performance of our method compared to the state of the art baseline models. Note here that the trade off afforded by our method is the ability to (1) generate much more finegrained explanations (as shown in Figure~\ref{fig:explain}) and (2) extend the RLS extraction model to a variety of tasks like contradiction detection and abductive reasoning, while marginally losing out on overall inference accuracy.

\begin{table}[t]
\centering
\begin{tabular}{c|c|ccc}
\hline
D & \#qns & Ours\footnotemark[5] & ProofWriter & PRover \\ \hline
0         & 6299  & 100        & 100         & 100    \\
1         & 4434  & 100        & 99.1        & 99.0   \\
2         & 2915  & 100        & 98.6        & 98.8   \\
3         & 2396  & 100        & 98.5        & 99.1   \\
4         & 2134  & 100        & 98.7        & 98.8   \\
5         & 2003  & 100        & 99.3        & 99.3   \\ \hline
All       & 20192 & 100        & 99.2        & 99.3   \\ \hline
\end{tabular}
\caption{\label{table:depth5} Results of the different systems on D5 dataset. We observe that our method comfortably outperforms the baseline methods even when our system was only trained on sentences from D3 while the other two systems were trained on the D5 dataset itself.}
\end{table}
\subsubsection{Generalization to higher depths}
We argue in the introduction that our method can perform reasoning to arbitrary depths and it is clear that the performance of a method which uses a symbolic reasoning engine is not constrained by the depths of reasoning. However, to demonstrate the same in practice, we train a model on the sentences from the D3 dataset and test it on the D5 dataset where the reasoning depths vary from 0 to 5. The results for the experiment are shown in Table~\ref{table:depth5} where we observe that our method registers perfect performance (100\% accuracy), reinforcing the advantages offered by our system. Furthermore, we observe perfect representation extraction (exact match) accuracy on the test set which highlights how the system can be easily extended to accurately perform the abduction and implication enumeration tasks which were defined in \citealt{proofwriter}, since they use a similar set of natural language templates as the D5 dataset.


\subsubsection{Zero shot generalization}
\begin{table}[t]
\centering
\begin{tabular}{c|c|ccc}
\hline
Test   & \#qns & Ours & ProofWriter & PRover \\ \hline
Birds1 & 40    & 90.0       & 95.0        & 95.0   \\
Elec4  & 4224  & 100        & 97.1        & 84.8   \\ \hline
All    & 5270  & 99.8       & 97.0        & 86.5   \\ \hline
\end{tabular}
\caption{\label{table:birds} Zero-shot results of models trained on some version of \ruletaker{} (D3 for our method and ProofWriter and D5 for PRover) and tested on the hand authored rulebases of the Birds-Electricity dataset. The results show how our model performs better than the baseline methods, demonstrating the generalization capability of our approach to out-of-distribution test sets.}
\end{table}

\begin{figure}[t]
\centering
\includegraphics[width=4cm,height=2.8cm]{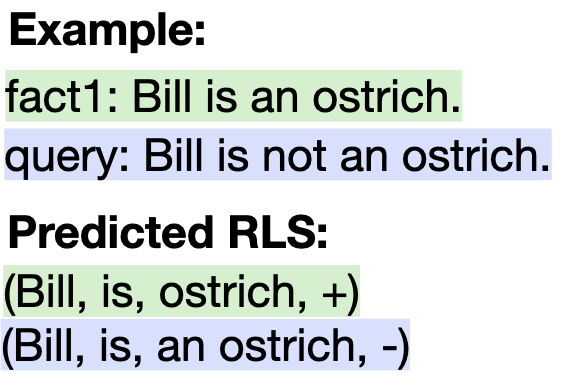}
\caption{\label{fig:birds} An example mistake of our method on the Birds dataset, highlighting a shortcoming of our current approach. However, we note that this specific mistake can be addressed by using a smarter unification operator as part of the symbolic reasoner which can determine whether two phrases are effectively equivalent.
}
\end{figure}

To demonstrate the generalization ability of our deductive reasoning method to unseen, out-of-distribution test sets we test the model trained on the D3 sentences (from Table~\ref{table:depth5}) in a zero shot setting on the hand authored rulebases of the Birds-Electricity dataset. The results show how our method performs competitively to the baseline models thus demonstrating the out-of-distribution generalization ability of our method. For further analysis, we show an example from the Birds dataset in Figure~\ref{fig:birds} where our method fails. The example highlights a shortcoming of the method and emphasizes the need for a smarter unification operator (as opposed to the default exact matching operator) as part of the symbolic reasoner, which can effectively judge the similarity of two phrases.


\subsection{Results summary}
\footnotetext[5]{see footnote 4.}
In summary, we observe that the answer to our questions regarding the generalization ability of the representation extraction model and the reasoning model is encouragingly positive. This supports our proposal of identifying and extracting RLS from natural language arguments as a noteworthy alternative to solving the natural language reasoning tasks, especially given the numerous advantages of the approach as discussed in Section~\ref{sec:advantages}.



\section{Related work}
This work builds on a recent line of work \cite{proofwriter, saha-etal-2020-prover, ruletaker} that tries to tackle the problem of natural language reasoning. The recent works make use of a transformer based model to solve the rule based reasoning task in an end-to-end manner while simultaneously predicting some form of explanations for the model's answer. However, in this work we move away from the end-to-end methods and instead propose identifying and extracting structured representations from natural language arguments and using a symbolic reasoning engine to solve the reasoning problem. This approach is inspired by numerous previous attempts \cite{Hayes1977InDO,HSME88,McCarthy63,McCarthy76} at similar tasks which proposed parsing natural language utterances into different structured meaning representations. However, in this work we only try to extract the logical structure from natural language arguments instead, which we argue is a reasonable assumption. 


\section{Discussion and Conclusion}
This work largely relies on our initial argument that it is possible to extract the logical structure from a NL argument. The community mostly agrees that NL is very expressive and cannot be mapped into structured representations; however, we believe that this does not contradict our underlying assumptions that ``locally" logical arguments made in language can be mapped to their core logical structure. After all, this is how humans discuss these arguments, and communicate about it -- they identify the core arguments and atoms, and use them to develop explanations. This is the heart of the argument we make in this paper. And, as we show, the mapping process can be accomplished at high accuracies, rivaling end-to-end methods, while providing more reasoning and explanation capabilities.

In conclusion, we began this work by noting the importance of finer explanations for natural language reasoning tasks, which is missing in current methods that only provide simple sentence level explanations. To address this and the lack of extensibility in current systems, we proposed identifying and extracting intermediate logical representations, and reasoning over the same as an approach to the task. Consequently, we note that previous works which tried to parse NL sentences into logical representations have understandably failed. However, we instead argue that it is possible to extract the core logical structure from a natural language argument and towards that end, we formulate Representations of Logical Structure (RLS) as a class of structured representations defined in the context of a reasoning task. We further show how our RLS extraction system can generalize to natural forms of language and different RLS formulations by showing its strong performance (95\% and 99\% exact match accuracy) on the datasets of \clutrr{} and \lot{} respectively. We also show the efficacy of our reasoning method on the deductive reasoning task on a series of \ruletaker{} datasets. 
We motivate future work to further explore methods to extract more contextual structured representations and extend it to support wider forms of natural language utterances.

\bibliography{bib/anthology,bib/custom,bib/ccg,bib/ccg_cited}
\bibliographystyle{bib/acl_natbib}




\end{document}